\documentclass[final,12pt]{msml2022} % final submission
\title[Momentum Transformer]{Momentum Transformer: Closing the Performance Gap Between Self-attention and Its Linearization}
\usepackage{times}
\usepackage{amsmath}
\usepackage{amssymb}
\usepackage{bbm}
\usepackage{cleveref}
\usepackage{multirow}
\usepackage{hhline}
\usepackage{natbib}
\usepackage{algcompatible}
\usepackage{algorithm}
\usepackage{enumitem}
\usepackage{multicol}
\usepackage{bm}
\usepackage{svg}
\usepackage{titlesec}
\usepackage{booktabs}       % professional-quality tables

\def\vb{{\bm{b}}}

\def\vk{{\bm{k}}}

\def\vm{{\bm{m}}}

\def\vp{{\bm{p}}}
\def\vq{{\bm{q}}}

\def\vs{{\bm{s}}}

\def\vu{{\bm{u}}}
\def\vv{{\bm{v}}}

\def\vx{{\bm{x}}}
\def\vy{{\bm{y}}}
\def\vz{{\bm{z}}}

\def\mA{{\bm{A}}}

\def\mI{{\bm{I}}}

\def\mK{{\bm{K}}}

\def\mQ{{\bm{Q}}}

\def\mT{{\bm{T}}}

\def\mV{{\bm{V}}}
\def\mW{{\bm{W}}}
\def\mX{{\bm{X}}}

\def \Ib {{\mathbf{Ib}}}

\def \RR {{\mathbb{R}}}

\def \Ib {{\mathbf{I}}}

\usepackage{verbatim}

\usepackage{wrapfig}

\msmlauthor{%
 \Name{Tan Nguyen} \Email{mn15@rice.edu}\\
 \addr Department of Mathematics, UCLA, Los Angeles, CA, 90095
 \AND
 \Name{Richard G. Baraniuk} \Email{richb@rice.edu}\\
 \addr Department of ECE, Rice University, Houston, TX, 77005
 \AND
 \Name{Robert M. Kirby} \Email{kirby@sci.utah.edu}\\
 \addr School of Computing and Scientific Computing and Imaging Institute, University of Utah, Salt Lake City, UT, 84112
 \AND
 \Name{Stanley J. Osher} \Email{sjo@math.ucla.edu}\\
 \addr Department of Mathematics, UCLA, Los Angeles, CA, 90095
 \AND
 \Name{Bao Wang} \Email{wangbaonj@gmail.com}\\
 \addr Department of Mathematics and Scientific Computing and Imaging Institute, University of Utah, Salt Lake City, UT, 84112
}

\makeatletter
 \let\Ginclude@graphics\@org@Ginclude@graphics
\makeatother

\begin{document}

\maketitle

\begin{abstract}%
Transformers have achieved remarkable success in sequence modeling and beyond but suffer from quadratic computational and memory complexities with respect to the length of the input sequence. Leveraging techniques include sparse and linear attention and hashing tricks; efficient transformers have been proposed to reduce the quadratic complexity of transformers but significantly degrade the accuracy. In response, we first interpret the linear attention and residual connections in computing the attention map as gradient descent steps. We then introduce momentum into these components and propose the \emph{momentum transformer}, which utilizes momentum to improve the accuracy of linear transformers while maintaining linear memory and computational complexities. Furthermore, we develop an adaptive strategy to compute the momentum value for our model based on the optimal momentum for quadratic optimization. This adaptive momentum eliminates the need to search for the optimal momentum value and further enhances the performance of the momentum transformer. A range of experiments on both autoregressive and non-autoregressive tasks, including image generation and machine translation, demonstrate that the momentum transformer outperforms popular linear transformers in training efficiency and accuracy.
\end{abstract}

\begin{keywords}%
  transformer, adaptive momentum, efficient attention
\end{keywords}

%\vspace{-0.3cm}
\section{Introduction}\label{sec:intro}%\vspace{-0.3cm}
The self-attention mechanism is the backbone of 
%plays a fundamental role in 
building transformers 
\citep{vaswani2017attention,kim2017structured}. 
Given an input sequence ${\mX}=[\vx_1,\ldots,\vx_N]^\top\in \RR^{N\times D_x}$ of $N$ feature vectors, the self-attention transforms it into another sequence $\hat{\mV}=[\hat{\vv}_1,\ldots,\hat{\vv}_N]^\top\in \RR^{N\times D_v}$ as follows
%which transforms the input sequence ${\mX}=[\vx_1,\vx_2,\ldots,\vx_N]^\top\in \RR^{N\times D_x}$ of $N$ feature vectors, each of which is encoded in a $D_x$-dimensional space, into another sequence $\hat{\mV}=[\hat{\vv}_1,\hat{\vv}_2,\ldots,\hat{\vv}_N]^\top\in \RR^{N\times D_v}$ as follows
\begin{equation}\label{eq:attention-single-vector}%\small
\hat{\vv}_i=\sum_{j=1}^N{\rm softmax}\Big(\frac{\vq_i^\top\vk_j}{\sqrt{D}} \Big)\vv_j,\ \mbox{for}\ i=1,\ldots,N,
\end{equation}
where the scalar ${\rm softmax}({(\vq_i^\top\vk_j)}/{\sqrt{D}})$ can be understood as the attention $\hat{\vv}_i$ pays to the input feature $\vx_j$. The vectors $\vq_i,\vk_j,$ and $\vv_j$ are %called 
the query, key, and value vectors, respectively, and are %these vectors are 
computed as follows
\begin{equation}\label{eq:query-key-value}%\small
\begin{aligned}
[] [\vq_1,\vq_2,\ldots,\vq_N]^\top &:= {\mQ} = {\mX}{\mW}_Q^\top \in \RR^{N\times D}, \\
[\vk_1,\vk_2,\ldots,\vk_N]^\top &:={\mK} = {\mX}{\mW}_K^\top \in \RR^{N \times D}, \\
[\vv_1,\vv_2,\ldots,\vv_N]^\top &:= {\mV} = {\mX}{\mW}_V^\top \in \RR^{N\times D_v}, 
\end{aligned}
\end{equation}
where ${\mW}_Q, {\mW}_K\in \RR^{D\times D_x}$, and ${\mW}_V\in \RR^{D_v\times D_x}$ are the weight matrices. %Note that
\eqref{eq:attention-single-vector} can be written as
%We can further write \eqref{eq:attention-single-vector} 
%as follows
%into the following compact form
\begin{equation}\label{eq:attention-vector-form}%\small
\hat{\mV}={\rm softmax}\Big(\frac{{\mQ}{\mK}^\top}{\sqrt{D}} \Big){\mV},
\end{equation}
where the softmax function is applied to each row of the matrix ${%\small 
{(\mQ\mK^\top)}/{\sqrt{D}}}$. %Equation
\eqref{eq:attention-vector-form} is also called the 
%``scaled dot-product attention'' or 
``softmax attention''. Each transformer layer $T_{\ell}(\cdot)$ is defined via the following residual connection,
\begin{align}%\small
    T_{\ell}({\mX}) = f_{\ell}(\hat{\mV} + {\mX}), \label{eqn:res-connect}
\end{align}
where $f_{\ell}(\cdot)$ is a function that transforms each feature vector independently and usually chosen to be a %small 
feedforward network. In this paper, we call a transformer built with softmax attention standard transformer or transformer. It is easy to see that both memory and computational complexity of \eqref{eq:attention-vector-form} are 
$\mathcal{O}(N^2)$ with $N$ being the length of the input sequence. We can further introduce causal masking into \eqref{eq:attention-vector-form} for autoregressive applications \citep{vaswani2017attention}.

Transformers have become the state-of-the-art model for solving many challenging problems in natural language processing
\citep{vaswani2017attention,47866,dai2019transformer,williams-etal-2018-broad,devlin2018bert,NEURIPS2020_1457c0d6,howard-ruder-2018-universal,rajpurkar-etal-2016-squad} and computer vision \citep{dehghani2018universal,so2019evolved,dosovitskiy2020image,touvron2020deit}. Nevertheless, the quadratic memory and computational cost of computing the softmax attention \eqref{eq:attention-vector-form} is a major bottleneck for applying transformers to large-scale applications that involve very long sequences, such as those in 
\citep{liu2018generating,huang2018music,pmlr-v80-parmar18a}.
Thus, much recent research on transformers has been focusing on developing efficient transformers, aiming to reduce the memory and computational complexities of the model \citep{qiu2019blockwise,child2019generating,ho2019axial,pmlr-v80-parmar18a,beltagy2020longformer,ainslie-etal-2020-etc,wang2020linformer,tay2020synthesizer,pmlr-v119-tay20a,Kitaev2020Reformer,roy-etal-2021-efficient,vyas2020fast,zaheer2021big,wang2020linformer,katharopoulos2020transformers,choromanski2021rethinking,shen2021efficient,DBLP:journals/corr/abs-2102-11174,pmlr-v80-blanc18a,NEURIPS2019_e43739bb,song2021implicit,peng2021random,xiong2021nystromformer,nguyen2021fmmformer}. A thorough survey of recent advances in efficient transformers is available at \citep{tay2020efficient}. 
These efficient transformers have better memory and/or computational efficiency at the cost of a significant reduction in accuracy.

\subsection{Motivation}
\citet{katharopoulos2020transformers} have established a connection between transformers and recurrent neural networks (RNNs) through the kernel trick. They propose the linear transformer, which can be considered a rank-one approximation of the softmax transformer. {Linear transformers have computational advantages in training, test, and inference: the RNN formulation (see Equation~\eqref{eq:rnn:formulation} below) enjoys fast inference, especially for autoregressive tasks, and the unrolled RNN formulation (see Equation~\eqref{eq:attention-kernel-linear-matrix} below) is efficient for fast training.} See section~\ref{sec:transformers}
%2 
for a detailed review of the linear transformer and its advantages. {\citet{MomentumRNN} proposes integrating momentum into RNNs to accelerate training RNNs and improve learning long-term dependencies. We notice that MomentumRNN also enjoys a closed unrolling form, which is quite unique among existing techniques for improving RNNs, enabling fast training, test, and inference; see section~\ref{sec:momentum-transformers}
%3 
for details.} As such, in this paper we study \emph{how does momentum improves linear transformers?}

% Mention the efficiency. 

%\BW{I. Momentum has been shown to accelerate training and improve test accuracy of RNNs in [31], which directly motivates us to leverage momentum to improve the RNN formulation of the linearized transformer.II. One major challenge of incorporating techniques that can improve RNNs to improve transformers lies in that equations (6) or (7) are used for parallel processing the input of linear transformers, while the equivalent RNN form in equation (8) is used for efficient autoregressive inference. For the improved RNNs, we need to find their closed unrolling form like in equations (6) or (7), or something that can be computed effectively. The momentum RNN is a good choice since its unrolling form can be computed explicitly, see equation (13). III. Our initial trial showed that incorporating momentum into the linear attention unit can improve performance, which motivated us to consider replacing skip connections in transformers with the proposed adaptive momentum connection. We have added motivation in our revised manuscript to make the paper more readable.}

%\BW{Why momentum is complementary to linear transformer?}

\subsection{Contribution}

%\BW{Mention computational efficiency.}

We propose \emph{momentum transformers} by integrating two new momentum-related ingredients into the recently proposed linear transformers \citep{katharopoulos2020transformers} and its RNN formulation to improve the model's accuracy and efficiency. Our contributions include: 1) Similar to \citep{MomentumRNN}, we make an analogy between the RNN formulation of causal linear attention, which is the linear attention with causal masking for auto-regressive applications \citep{katharopoulos2020transformers}, and a gradient descent step. We then integrate the heavy ball-style momentum into this RNN formulation and result in the causal momentum attention. We extend this causal momentum attention into momentum attention for both autoregressive and non-autoregressive applications. We name the transformer with the new momentum attention the \emph{momentum transformer}. 2) We further introduce another momentum into the residual connection between the attention $\hat{\mV}$ and the input ${\mX}$ in~\eqref{eqn:res-connect} to enhance the model's performance. 3) We develop a new adaptive strategy to compute the momentum value and eliminate the burden of tuning momentum hyperparameters in our model. %Our adaptive momentum is principled and derived from the theoretical optimal choice of the heavy ball momentum for quadratic optimization problems. 
We name the momentum transformer with adaptive momentum the \emph{adaptive momentum transformer}.
The major advantages of momentum-based transformers include: 
%our momentum-based transformers include: 
\begin{itemize}%\vspace{-0.25cm}
\item Momentum and adaptive momentum transformers inherit memory and computational efficiency from the linear transformers while achieving better accuracy. %Furthermore, the RNN formulation of our models with causal masking enables fast inference for autoregressive applications. 

%\vspace{-0.25cm}
\item The training of momentum-based transformers converges remarkably faster than the training of linear transformers.

%\vspace{-0.25cm}
\item The design principle of momentum transformers is rooted in momentum-based optimization algorithms, enabling us to design more general and advanced momentum transformers for a wide range of applications.
\end{itemize}

\subsection{Related Work}\label{subsec:related-work}
In this part, we briefly review three lines of recent research that are most related to our work: 1) momentum in optimization and sampling, 2) momentum in deep neural network (DNN) design and 3) algorithms for efficient transformers.
%efficient transformers with linear memory and computational complexities.

\paragraph{Momentum in optimization and sampling} Momentum has been a popular technique for accelerating (stochastic) gradient-based optimization \citep{polyak1964some,goh2017momentum,sutskever2013importance,kingma2014adam,%,bengio2013advances,
paszke2019pytorch,sun2021training} and sampling algorithms \citep{duane1987hybrid,neal2011mcmc,chen2014stochastic,betancourt2017conceptual}. A particularly interesting momentum is the iteration-dependent one in NAG \citep{nesterov1983method,nemirovskii1985optimal,beck2009fast}, which has a significantly better convergence rate than constant momentum for convex optimization. The stochastic gradient NAG that employs a scheduled restart can also be used to accelerate DNN training with better accuracy and faster convergence \citep{wang2020scheduled}.

\paragraph{DNNs with momentum.} Momentum has been used for designing DNN architectures. \citet{he2019momentum} employ momentum to build large and consistent dictionaries for unsupervised learning with a contrastive loss leveraging
%. At the core of this approach is a 
momentum-based moving average of the queue encoder. Many DNN-based methods for sparse coding are designed by unfolding the classical optimization algorithms with momentum,  %\citep{moreau2017understanding}, 
e.g., FISTA \citep{beck2009fast}.
%, in which momentum can be used in the underpinning optimizer. 
MomentumRNNs \citep{MomentumRNN,wang2021does} are a class of RNNs that are designed based on momentum accelerated first-order optimization algorithms. MomentumRNNs can effectively resolve the vanishing gradient issue in training RNNs and obtain faster training and better performance over traditional RNNs. Momentum has also been integrated into ResNets \citep{li2018optimization} and neural ODEs \cite{NEURIPS2021_9a86d531}.
%,sander2021momentum}.

\paragraph{Efficient transformers.}
Existing efficient transformers can be roughly classified into several categories, as summarized in \citep{roy-etal-2021-efficient}. Among these categories are models with fixed patterns, which sparsify the attention matrix  \citep{pmlr-v80-parmar18a,liu2018generating,qiu2019blockwise,child2019generating,beltagy2020longformer}. Another category includes models that integrate two or more distinct access patterns to improve the coverage  \citep{child2019generating,ho2019axial}. Learnable patterns are also leveraged to learn the access pattern in a data-driven fashion \citep{Kitaev2020Reformer,roy-etal-2021-efficient,pmlr-v119-tay20a}. Some other efficient transformers take advantage of a side memory module to access multiple tokens at once \citep{lee2019set,sukhbaatar2019augmenting,asai2020challenges,beltagy2020longformer}. Finally, low-rank and kernelization approximation are employed to improve the memory and computational efficiency of computing self-attention, see e.g., \citep{tsai2019transformer,wang2020linformer,katharopoulos2020transformers,choromanski2021rethinking,shen2021efficient}.

\subsection{Notations}
We denote scalars by lower- or upper-case letters. We also denote vectors and matrices by lower- and upper-case boldface letters, respectively. For a vector $\vx = (x_1, \ldots, x_d)^\top\in \mathbb{R}^d$, where $(x_1,\ldots,x_d)^\top$ denotes the transpose of the vector $(x_1,\ldots,x_d)$, we use $\|\vx\| = {(\sum_{i=1}^d |x_i|^2)^{1/2}}$ to denote its $\ell_2$ norm. We denote the vector whose entries are all 0s as $\mathbf{0}$. For a matrix $\mA$, we use $\mA^\top$,  $\mA^{-1}$, and $\|\mA\|$ to denote its transpose, inverse, and spectral norm, respectively. We use $\mI$ to denote the identity matrix, whose dimension can be determined in its context.
For a function $f(\vx): \mathbb{R}^d \rightarrow \mathbb{R}$, we denote its gradient as $\nabla f(\vx)$. Given two sequences $\{a_n\}$ and $\{b_n\}$, we write $a_n=\mathcal{O}(b_n)$ if there exists a positive constant $0<C<+\infty$ such that $a_n \leq C b_n$.

\subsection{Organization}
We organize this paper as follows: In section~\ref{sec:transformers}, 
we review the kernelization trick used to linearize the softmax attention and the RNN formulation of the linear transformer with causal masking. In section~\ref{sec:momentum-transformers}, 
we present the momentum transformer and adaptive momentum transformer, providing the motivation and detailed derivation. We verify the efficiency of our momentum-based transformers on various applications, including both autoregressive and non-autoregressive tasks in section~\ref{sec:experiments}. 
The paper ends up with concluding remarks.

\section{Linear Transformer}
\label{sec:transformers}
Transformers learn long-term dependencies in %long 
sequences effectively and concurrently through the self-attention mechanism. %\eqref{eq:attention-single-vector}. 
By denoting $k(\vq_i,\vk_j)$ $:=\exp(\vq_i^\top\vk_j/\sqrt{D})$, we can rewrite \eqref{eq:attention-single-vector} as 
\begin{align}
{\hat{\vv}_i=({\sum_{j=1}^Nk(\vq_i,\vk_j)\vv_j})/({\sum_{j=1}^Nk(\vq_i,\vk_j)})}. \nonumber
\end{align}
In linear transformers \citep{wang2020linformer,katharopoulos2020transformers,choromanski2021rethinking,shen2021efficient}, the feature map $k(\vq_i,\vk_j)$ is linearized as the product of feature maps $\phi(\cdot)$ on the vectors $\vq_i$ and $\vk_j$, i.e., $k(\vq_i,\vk_j)=\phi(\vq_i)^\top\phi(\vk_j)$. The associative property of matrix multiplication is then utilized to derive the following efficient computation of the attention map
\begin{equation}
\begin{aligned}\label{eq:attention3}
\hat{\vv}_i=\frac{\sum_{j=1}^Nk(\vq_i,\vk_j)\vv_j}{\sum_{j=1}^Nk(\vq_i,\vk_j)}
=\frac{\sum_{j=1}^N\phi(\vq_i)^\top\phi(\vk_j)\vv_j}{\sum_{j=1}^N\phi(\vq_i)^\top\phi(\vk_j) } =\frac{\phi(\vq_i)^\top\sum_{j=1}^N\phi(\vk_j)\vv_j^\top}{\phi(\vq_i)^\top\sum_{j=1}^N\phi(\vk_j)}.
\end{aligned}
\end{equation}
In the matrix-product form, we can further write \eqref{eq:attention3} as follows
\begin{equation}\label{eq:attention-kernel-linear-matrix}
\Hat{\mV}=\frac{\phi({\mQ})(\phi({\mK})^\top{\mV} )}{\phi({\mQ})\phi({\mK})^\top}.
\end{equation}
%Note that 
Replacing $(\phi({\mQ})\phi({\mK}^\top)){\mV}$ with $\phi({\mQ})(\phi({\mK}^\top){\mV})$ reduces the  memory and computational cost of computing the attention map from $\mathcal{O}(N^2)$ to $\mathcal{O}(N)$, making linear transformers scalable to very long sequences. 

Furthermore, causal masking can be easily implemented in the linearized attention by truncating the summation term in the last equation of \eqref{eq:attention3}, resulting in 
\begin{eqnarray}\label{eq:attention:causal-masking}
\hat{\vv}_i=\frac{\phi(\vq_i)^\top\sum_{j=1}^i\phi(\vk_j)\vv_j^\top}{\phi(\vq_i)^\top\sum_{j=1}^i\phi(\vk_j)}:=\frac{\phi(\vq_i)^\top\vs_i}{\phi(\vq_i)^\top\vz_i},
\end{eqnarray}
where $\vs_i=\sum_{j=1}^i\phi(\vk_j)\vv_j^\top$ and $\vz_i=\sum_{j=1}^i\phi(\vk_j)$. The states $\vs_i$ and $\vz_i$ can be computed in a recurrent fashion.

%recurrently.

\paragraph{Efficient inference via the RNN formulation.} Self-attention processes tokens of a sequence concurrently, enabling fast training of transformers via layerwise parallelism. However, during inference, the output for timestep $i$ is the input for timestep $i + 1$. As a result, the inference in standard transformers cannot be parallelized and is thus computationally inefficient. Linear transformers provide an elegant approach to fixing this issue by leveraging their RNN formulation. In particular, we can further write the linear attention with causal masking in \eqref{eq:attention:causal-masking} into the following RNN form\footnote{For simplicity, we omit the nonlinearity (a two-layer feedforward network) compared to \citep{katharopoulos2020transformers}.}
\begin{equation}\label{eq:rnn:formulation}
\begin{aligned}
\vs_i&=\vs_{i-1} + \phi(\vk_i)\vv_i^\top;\quad\\
\vz_i&=\vz_{i-1} + \phi(\vk_i);\quad\\
\hat{\vv}_i&=\frac{\phi(\vq_i)^\top \vs_i}{\phi(\vq_i)^\top\vz_i},
\end{aligned}
\end{equation}
where $\vs_0=\mathbf{0}$ and $\vz_0=\mathbf{0}$. Note that this RNN formulation of linear transformers with causal masking contains two memory states $\vs_i$ and $\vz_i$.

\section{Momentum Transformer}
\label{sec:momentum-transformers}
In this section, we present the \emph{momentum transformer}. We start by integrating the heavy ball momentum into the RNN formulation of causal linear attention in \eqref{eq:rnn:formulation}, resulting in the causal momentum attention. Next, we generalize the causal momentum attention to momentum attention that can efficiently train the model. Moreover, we propose the \emph{momentum connection} to replace residual connections between the attention $\hat{\mV}$ and the input ${\mX}$ in~\eqref{eqn:res-connect} to boost the model's performance. Finally, we derive the adaptive momentum attention from the theory of optimal choice of momentum for the heavy ball method.

\subsection{Momentum Transformer}\label{subsec:momentum-transformer}
\paragraph{Heavy ball momentum.} Let us recall the heavy ball momentum for accelerating gradient descent in solving $\min_{\vx\in \RR^d}f(\vx)$ \citep{polyak1964some}. Starting from $\vx^0$ and $\vx^1$, the heavy ball method iterates as follows
\begin{equation}\label{eq:HB1}
\vx^{k+1} = \vx^k-\gamma\nabla f(\vx^k) + \beta(\vx^k-\vx^{k-1}),
\end{equation}
where $\gamma>0$ is the step size and $0\leq \beta <1$ is the momentum parameter. By introducing the momentum state $\vm$, we can rewrite the HB method as
\begin{equation}\label{eq:HB2}
\begin{aligned}
{\vm}^{k+1} = \beta{\vm}^k+\nabla f(\vx^k);\quad
{\vx}^{k+1} = {\vx}^k-\gamma{\vm}^{k+1}.
\end{aligned}
\end{equation}
In contrast, gradient descent updates at each step as follows
\begin{equation}\label{eq:GD}
\vx^{k+1} = \vx^k-\gamma \nabla f(\vx^k).
\end{equation}
%Heavy ball momentum can accelerate gradient descent and stochastic gradient descent \citep{sutskever2013importance}. 

\paragraph{Integrating momentum into causal linear attention.} Now we consider integrating the heavy ball momentum into causal linear attention. We integrate momentum into the state $\vs_i$  in \eqref{eq:rnn:formulation} only since the denominator in causal linear attention is simply a normalizing scalar. If we regard $-\phi(\vk_i)\vv_i^\top$ as the gradient vector in \eqref{eq:GD}, then we can add momentum into the state $\vs_i$ by following %the principle of 
the heavy ball method in \eqref{eq:HB2}, resulting in the following RNN formulation of causal momentum attention,
\begin{equation}\label{eq:momentum:rnn:formulation}
\begin{aligned}
\vm_i &= \beta\vm_{i-1}-\phi(\vk_i)\vv_i^\top;\quad \\
\vs_i &= \vs_{i-1} - \gamma\vm_i;\quad\\
\vz_i &= \vz_{i-1} + \phi(\vk_i);\quad\\
\hat{\vv}_i &= \frac{\phi(\vq_i)^\top\vs_i}{\phi(\vq_i)^\top\vz_i},
\end{aligned}
\end{equation}
where $\vm_0=\mathbf{0}$, and $\gamma>0$ and $0\leq \beta<1$ are two hyperparameters. The RNN formulation of causal momentum attention in \eqref{eq:momentum:rnn:formulation} is efficient for autoregressive inference. For efficient training, we need to rewrite \eqref{eq:momentum:rnn:formulation} into a form that is similar to the linear attention in \eqref{eq:attention:causal-masking}. To this end, we need to eliminate the states $\vm_i, \vs_i$, and $\vz_i$ from \eqref{eq:momentum:rnn:formulation}. Notice that
\begin{equation*}
\begin{aligned}
\vs_i = \vs_{i-1}-\underbrace{\gamma\vm_i}_{:=\vp_i}
 = \underbrace{\vs_0}_{=\mathbf{0}} - \Big(\vp_i+\vp_{i-1}+\ldots+\vp_1\Big),
\end{aligned}
\end{equation*}
since $\vm_i=\beta\vm_{i-1}-\phi(\vk_i)\vv_i^\top$, we have $\vp_i=\beta\vp_{i-1}-\gamma\phi(\vk_i)\vv_i^\top$. Therefore,
\begin{equation*}
\begin{aligned}
\vs_i &= - (\vp_i+\vp_{i-1}+\ldots+\vp_1)\\
&= \gamma\phi(\vk_i)\vv_i^\top - \Big( (1+\beta)\vp_{i-1}+\vp_{i-2}+\ldots+\vp_1\Big)\\
&= \gamma\phi(\vk_i)\vv_i^\top + \gamma(1+\beta)\phi(\vk_i)\vv_i^\top \\
&\ \ \ \ \ - \Big((1+\beta)^2\vp_{i-2}+\ldots+\vp_1\Big)\\
&=\ldots\\
&=\gamma \sum_{j=1}^i\frac{1-\beta^{i-j+1}}{1-\beta}\phi(\vk_j)\vv_j^\top\ \mbox{for}\ i\geq 1.
\end{aligned}
\end{equation*}

We can then formulate the causal momentum attention as follows
\begin{equation}\label{eq:momentum:causal:attention}
\hat{\vv}_i=\frac{\gamma\phi(\vq_i)^\top\sum_{j=1}^i\Big(\frac{1-\beta^{i-j+1}}{1-\beta}\phi(\vk_j)\vv_j^\top\Big) }{\phi(\vq_i)^\top\vz_i}.
\end{equation}
Note that \eqref{eq:momentum:causal:attention} is mathematically equivalent to \eqref{eq:momentum:rnn:formulation}, but it can be trained much more efficiently in a concurrent fashion via layerwise parallelism. 

\begin{remark}
Comparing \eqref{eq:momentum:causal:attention} with \eqref{eq:attention:causal-masking} , we see that momentum plays a role in reweighting the terms $\{\phi(\vk_j)\vv_j^\top\}_{j=1}^i$. It is interesting to note that this reweighting is opposite to that used for reweighting the local attention \citep{dai2019transformer}. It has also been noticed that low-rank attention can complement local attention, resulting in improved performance \citep{nguyen2021fmmformer}. Often local attention behaves quite differently from low-rank attention, and different reweighting can be beneficial. {One particular reweighting strategy is decomposing softmax attention into long and short-range components and using different weighting schemes for each part.
}
%\citept{nguyen2021fmmformer} uses low-rank attention to model far-field attention while uses local-attention to model near-field attention. 
We leave the study of reweighting local attention and low-rank attention differently as future work. 
%\BW{[We have added more discussion on momentum transformers and the weighting of softmax attention. Cite FMMformer and add more discussion.]}
\end{remark}

\paragraph{Integrating momentum into linear attention.} To obtain momentum attention without causal masking, we can simply take the sum from $1$ to $N$ instead of summing from $1$ to $i$. Therefore, we obtain the following momentum attention
\begin{equation}\label{eq:momentum:attention}
\hat{\vv}_i=\frac{\gamma\phi(\vq_i)^\top\sum_{j=1}^N\Big(\frac{1-\beta^{N-j+1}}{1-\beta}\phi(\vk_j)\vv_j^\top\Big) }{\phi(\vq_i)^\top\sum_{j=1}^N\phi(\vk_j)}.
\end{equation}

\paragraph{Memory and computational complexity.} It is clear that training momentum transformers have the same memory and computational complexities of $\mathcal{O}(N)$ as the training of linear transformers. For test and inference, momentum transformers also have the same memory and computational complexities as linear transformers. However, in the RNN form, momentum transformers require slightly more memory than linear transformers to store the extra momentum state $\vm_i$.

\subsection{Momentum Connection}\label{subsec:momentum-connection}
On top of the self-attention unit, each transformer layer has a residual connection between the self-attention output and the input as shown in~\eqref{eqn:res-connect}. We further integrate momentum into~\eqref{eqn:res-connect} and derive the momentum connection as follows

\begin{equation}\label{eq:momentum-transformer-layer}
T_\ell({\mX}) = f_\ell\big(\hat{\mV}+{\mX} + \Tilde{\beta}({\mX}-T_{\ell-1}({\mX}))\big),
\end{equation}
where $0\leq \Tilde{\beta}<1$ is a hyperparameter. 

\subsubsection{Adaptive Momentum}\label{subsubsec:adaptive-momentum}\ \ 
Our momentum transformer introduces additional hyperparameters $\gamma$ and $\beta$, as well as $\Tilde{\beta}$, compared to the linear transformer. In section~\ref{sec:experiments}, 
we show that $\gamma$ can be simply set to 1 in many experiments. However, tuning the momentum-related hyperparameters $\beta$ and $\Tilde{\beta}$
can introduce extra computational cost for training transformers. Moreover, using a constant momentum may not give us optimal performance. In this section, we will introduce an adaptive momentum formula for computing the momentum hyperparameter in momentum connection and thus eliminating the computational overhead for tuning $\Tilde{\beta}$. Here, the adaptive momentum does not apply to $\beta$ since it will break the closed unrolling form in \eqref{eq:momentum:causal:attention}. 
%those values.

%\subsubsection{3.2.2. Optimal Choice of Heavy Ball Momentum}~~
\paragraph{Optimal choice of heavy ball momentum.}
We motivate our adaptive momentum from the optimal choice of the heavy ball momentum for solving the following quadratic minimization problem
\begin{equation}\label{eq:quadratic}
\min_{\vx} f(\vx):= \frac{1}{2}\vx^\top\mA\vx +\vx^\top\vb.
\end{equation}
The following theorem gives the optimal choice of $\beta$ for \eqref{eq:HB2} to solve \eqref{eq:quadratic}.

\begin{theorem}[\cite{sun2021training}]\label{th1}
Let $f(\vx)$ be the quadratic function \eqref{eq:quadratic} with $\nabla f(\vx)= \mA\vx+\vb$, where $\mA$ is positive definite. Moreover, we denote the smallest ($\lambda_{\min}(\mA)$) and the largest eigenvalues ($\lambda_{\max}(\mA)$) of $\mA$ as $\nu$ and $L$, respectively. Given any fixed step size $\gamma\leq{1}/{L}$, the optimal choice for $\beta$ is ${\tilde{\beta}}=(1-\sqrt{\gamma\nu})^2$. In this case, the heavy ball method achieves a convergence rate of $$\|{\vx}^{k+1}-{\vx}^*\|\leq (1-\sqrt{\gamma\nu})\|{\vx}^{k}-{\vx}^*\|,$$
where $\vx^*$ is the minimum of the quadratic function \eqref{eq:quadratic}.
\end{theorem}

Theorem~\ref{th1} shows that the optimal momentum for the heavy ball method should be %scaled by 
$(1-\sqrt{\gamma\nu})^2$ if $\gamma\leq 1/L$. However, the smallest eigenvalue $\nu$ is usually unknown. Therefore, we consider constructing the sequence $\{\|\nabla f(\vx^k)-\nabla f(\vx^{k-1})\|/\|\vx^k-\vx^{k-1}\|\}_{k\geq 1}$ to approximate $\nu$. We have the following theoretical result to guarantee that $\|\nabla f(\vx^k)-\nabla f(\vx^{k-1})\|/\|{\vx}^k-{\vx}^{k-1}\|\rightarrow \nu$ as $k\rightarrow \infty$. 

\begin{proposition}[\cite{sun2021training}]\label{prop1}
Assume that conditions in Theorem \ref{th1} hold and $\{{\vx}^k\}_{k\geq 0}$ is generated by the heavy ball method \eqref{eq:HB2}. If $\gamma\leq {1}/{L}$, for any fixed $0\leq 
{%\tilde
{\beta}}<1$, we have 
$$ \lim_{k\rightarrow \infty}\frac{\|\nabla f(\vx^k)-\nabla f(\vx^{k-1})\|}{\|{\vx}^k-{\vx}^{k-1}\|}=\nu.$$
\end{proposition}

%We will prove Theorem~\ref{th1} and Proposition~\ref{prop1} in Appendix~\ref{Appendix:proof}. 
In practice, for a given step size $\gamma$, we restrict the adaptive momentum to be in the range $[0,1-\delta]$ with $\delta$ being the threshold parameter, and we choose it to be $10^{-3}$ in this work. Hence, we have the following adaptive momentum
\begin{equation}
\label{eq:adp:momentum}
{\bf proj}_{[0,1-\delta]}\left(1-\sqrt{\gamma\frac{\|\nabla f(\vx^k)-\nabla f(\vx^{k-1})\|}{\|\vx^k-\vx^{k-1}\|}}\right)^2,
\end{equation}
where ${\bf proj}_{[0,1-\delta]}(\cdot):=\max(0,\min(1-\delta,\cdot))$. To simplify our computation, we apply the gradient descent update to approximate $\vx^k-\vx^{k-1}$, i.e., we approximate $\vx^k-\vx^{k-1}$ by $\gamma\nabla f(\vx^{k-1})$, and we end up with
\begin{equation}
\label{eq:adp:momentum2}
{%\small 
{\tilde{\beta}_k}:={\bf proj}_{[0,1-\delta]}\left(1-\sqrt{\frac{\|\nabla f(\vx^k)-\nabla f(\vx^{k-1})\|}{\|\nabla f(\vx^{k-1}) \|}}\right)^2.}
\end{equation}

%\label{sec:experiments}
\begin{figure*}[t!]
\centering
\includegraphics[width=\linewidth]{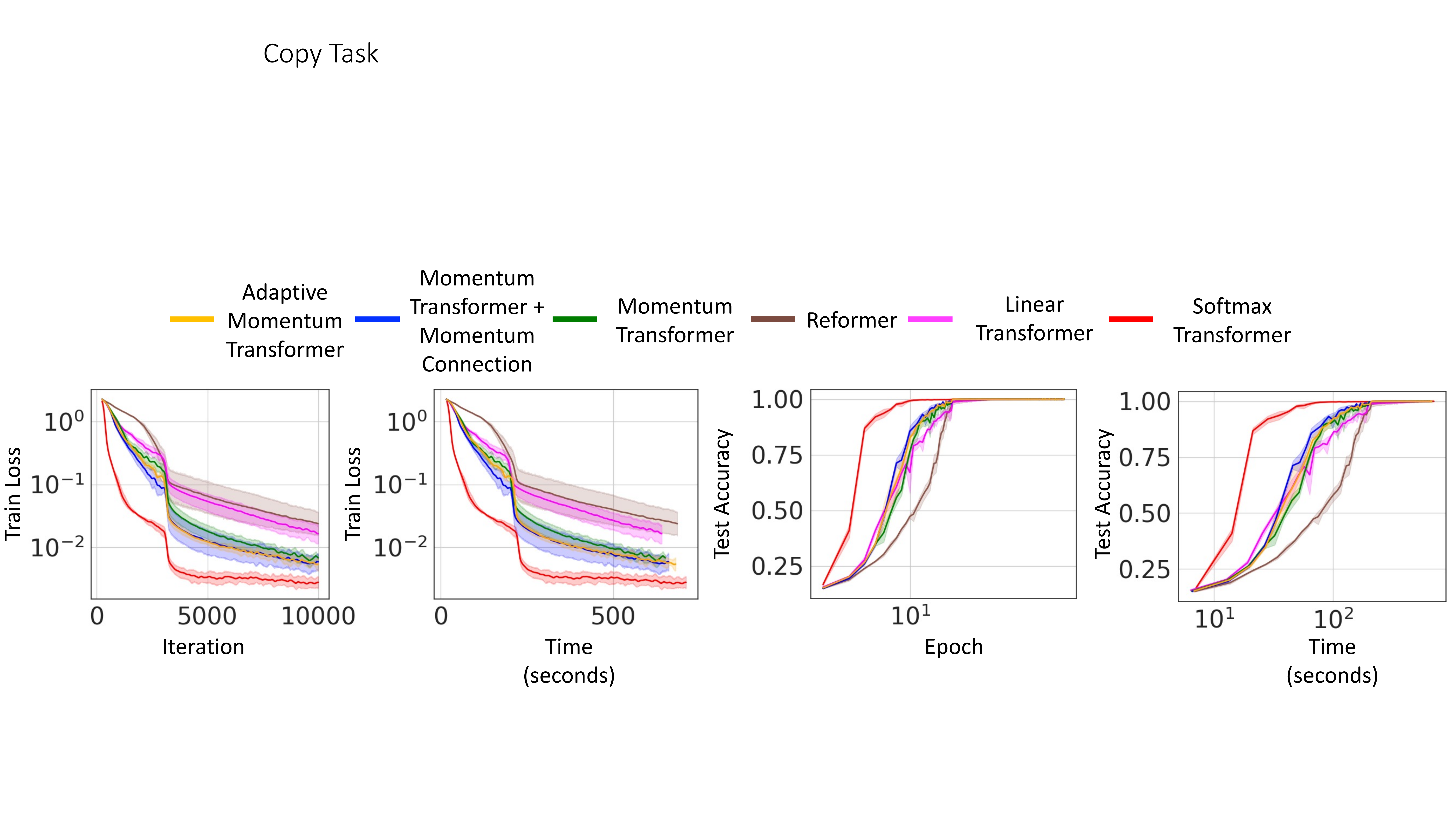}\vspace{-1cm}
\caption{Convergence comparison of adaptive momentum, momentum, reformer, linear, and softmax transformer on the sequence copy task. Momentum and adaptive momentum transformers converge faster and achieve better training loss than both linear transformer and reformer. Softmax transformer converges the fastest but suffers from quadratic memory and computational complexities. Adaptive momentum transformer performs as well as momentum transformer without intensively searching for momentum values. The training time per epoch for softmax transformer, reformer, linear transformer, our momentum transformer without and with momentum connection, and our adaptive momentum transformer are $7.1$, $6.9$, $6.4$, $6.5$, $6.6$, and $6.8$ seconds, respectively. Here, the computational time per epoch does not reveal a significant advantage of efficient transformers over the softmax transformer because the sequence length is only 128.
%\BW{[I believe a more informative: training time as the x-axis and the training loss as the y-axis. 
%How does momentum Transformer compare to softmax Transformer in that case?
%]} 
}
\label{fig:copy-convergence-analysis}
\end{figure*}

\section{Experimental Results}\label{sec:experiments}

In this section, we evaluate the benefits of our momentum transformers in terms of convergence speed, efficiency, and accuracy. We compare the performance of momentum and adaptive momentum transformers \footnote{{Here, the momentum and adaptive momentum transformers are built on the same baseline architecture of the linear transformer in terms of the number of layers and number of heads.}} with the baseline standard softmax transformer and several other efficient transformers in the following tasks: 1) the synthetic copy task, 2) the MNIST and CIFAR image generation task, 3) Long-Range Arena~\citep{tay2021long}, and 4) the non-autoregressive machine translation task. These tasks are among standard benchmarks for measuring the performance of transformers and their {computational and memory} efficiency. The tasks we choose also cover different data modalities --- text and image --- and a variety of model sizes. Our experimental results confirm that momentum and adaptive momentum transformers outperform many existing efficient transformers, including linear transformers and reformers, in accuracy and convergence.
%converge faster. 
Furthermore, {adaptive momentum transformer improves over momentum transformer without the need of searching for momentum hyperparameter for the momentum connection.}
%$\tilde{\beta}$.} 
% \subsection{Momentum and Stepsize Hyperparameteres for the Experiments}
Values of momentum-related hyperparameters for experiments in our experiments 
%and for experiments on the Long Range Arena benchmark in Section~\ref{appendix:lra} 
are provided in Table~\ref{tab:detailed-hyperparams} below.
\begin{table}[!t]
%\vspace{-0.1in}
    \caption{Momentum and Stepsize Hyperparameteres for Momentum-based Transformers.}
%\vspace{-0.2in}
\label{tab:detailed-hyperparams}
\begin{center}
{\footnotesize
\begin{tabular}{l|cccc}
    \toprule
    Model & Momentum & Stepsize & Momentum in & Stepsize in \\
     &  &  & Momentum Connection & Momentum Connection \\
    \midrule
    \midrule
    \multicolumn{5}{c}{Copy Task} \\
    \midrule
    % In MomentumLSTM, We set momentum to $0.6$ and step size to $0.6$ and $1.0$ for MNIST and PMNIST tasks, respectively. 
    % We set momentum $\mu$ and stepsize $\epsilon$ to 0.3 and 0.1, respectively.
% for srsgd, set rs to 2
% We set momentum $\mu$ and stepsize $\epsilon$ to 0.0 and 0.6, respectively.
    Momentum transformer & $0.1$ & $0.6$ & &  \\
    %\midrule
    Momentum transformer & $0.1$ & $0.6$ & $0.99$ & $0.99$ \\
    + Momentum connection &  &  &  & \\
    %\midrule
    Adatptive momentum transformer & $0.1$ & $0.6$ &  & $0.99$ \\
    \midrule
    \multicolumn{5}{c}{MNIST Generation} \\
    \midrule
    % In MomentumLSTM, We set momentum to $0.6$ and step size to $0.6$ and $1.0$ for MNIST and PMNIST tasks, respectively. 
    % We set momentum $\mu$ and stepsize $\epsilon$ to 0.3 and 0.1, respectively.
% for srsgd, set rs to 2
% We set momentum $\mu$ and stepsize $\epsilon$ to 0.0 and 0.6, respectively.
    Momentum transformer & $0.6$ & $0.9$ & &  \\
    %\midrule
    Momentum transformer  & $0.6$ & $0.9$ & $0.1$ & $0.99$ \\
    + Momentum connection &  &  &  & \\
    %\midrule
    Adatptive momentum transformer & $0.6$ & $0.9$ &  & $0.99$ \\
    \midrule
    \multicolumn{5}{c}{CIFAR Generation} \\
    \midrule
    % In MomentumLSTM, We set momentum to $0.6$ and step size to $0.6$ and $1.0$ for MNIST and PMNIST tasks, respectively. 
    % We set momentum $\mu$ and stepsize $\epsilon$ to 0.3 and 0.1, respectively.
% for srsgd, set rs to 2
% We set momentum $\mu$ and stepsize $\epsilon$ to 0.0 and 0.6, respectively.
    Momentum transformer & $0.1$ & $0.9$ & &  \\
    %\midrule
    Momentum transformer  & $0.1$ & $0.9$ & $0.1$ & $0.9$ \\
    + Momentum connection &  &  &  & \\
    %\midrule
    Adatptive momentum transformer & $0.1$ & $0.9$ &  & $0.9$ \\
    \midrule
    \multicolumn{5}{c}{Non-Autoregressive Machine Translation} \\
    \midrule
    % In MomentumLSTM, We set momentum to $0.6$ and step size to $0.6$ and $1.0$ for MNIST and PMNIST tasks, respectively. 
    % We set momentum $\mu$ and stepsize $\epsilon$ to 0.3 and 0.1, respectively.
% for srsgd, set rs to 2
% We set momentum $\mu$ and stepsize $\epsilon$ to 0.0 and 0.6, respectively.
    Momentum transformer & $0.6$ & $0.6$ & &  \\
    %\midrule
    Momentum transformer  & $0.6$ & $0.6$ & $0.3$ & $0.9$ \\
     + Momentum connection &  &  &  & \\
    %\midrule
    Adatptive momentum transformer & $0.6$ & $0.6$ &  & $0.9$ \\
    \midrule
    \multicolumn{5}{c}{ListOps} \\
    \midrule
    % In MomentumLSTM, We set momentum to $0.6$ and step size to $0.6$ and $1.0$ for MNIST and PMNIST tasks, respectively. 
    % We set momentum $\mu$ and stepsize $\epsilon$ to 0.3 and 0.1, respectively.
% for srsgd, set rs to 2
% We set momentum $\mu$ and stepsize $\epsilon$ to 0.0 and 0.6, respectively.
    Momentum transformer & $0.1$ & $0.6$ & &  \\
    %\midrule
    Adatptive momentum transformer & $0.1$ & $0.6$ &  & $0.4$ \\
    \midrule
    \multicolumn{5}{c}{Text} \\
    \midrule
    % In MomentumLSTM, We set momentum to $0.6$ and step size to $0.6$ and $1.0$ for MNIST and PMNIST tasks, respectively. 
    % We set momentum $\mu$ and stepsize $\epsilon$ to 0.3 and 0.1, respectively.
% for srsgd, set rs to 2
% We set momentum $\mu$ and stepsize $\epsilon$ to 0.0 and 0.6, respectively.
    Momentum transformer & $0.6$ & $2.0$ & &  \\
    %\midrule
    Adatptive momentum transformer & $0.6$ & $2.0$ &  & $0.001$ \\
    \midrule
    \multicolumn{5}{c}{Retrieval} \\
    \midrule
    % In MomentumLSTM, We set momentum to $0.6$ and step size to $0.6$ and $1.0$ for MNIST and PMNIST tasks, respectively. 
    % We set momentum $\mu$ and stepsize $\epsilon$ to 0.3 and 0.1, respectively.
% for srsgd, set rs to 2
% We set momentum $\mu$ and stepsize $\epsilon$ to 0.0 and 0.6, respectively.
    Momentum transformer & $0.6$ & $1.0$ & &  \\
    %\midrule
    Adatptive momentum transformer & $0.6$ & $1.0$ &  & $0.5$ \\
    \midrule
    \multicolumn{5}{c}{Image} \\
    \midrule
    % In MomentumLSTM, We set momentum to $0.6$ and step size to $0.6$ and $1.0$ for MNIST and PMNIST tasks, respectively. 
    % We set momentum $\mu$ and stepsize $\epsilon$ to 0.3 and 0.1, respectively.
% for srsgd, set rs to 2
% We set momentum $\mu$ and stepsize $\epsilon$ to 0.0 and 0.6, respectively.
    Momentum transformer & $0.9$ & $0.9$ & &  \\
    %\midrule
    Adatptive momentum transformer & $0.9$ & $0.9$ &  & $0.001$ \\
    \midrule
    \multicolumn{5}{c}{Pathfinder} \\
    \midrule
    % In MomentumLSTM, We set momentum to $0.6$ and step size to $0.6$ and $1.0$ for MNIST and PMNIST tasks, respectively. 
    % We set momentum $\mu$ and stepsize $\epsilon$ to 0.3 and 0.1, respectively.
% for srsgd, set rs to 2
% We set momentum $\mu$ and stepsize $\epsilon$ to 0.0 and 0.6, respectively.
    Momentum transformer & $0.3$ & $0.1$ & &  \\
    %\midrule
    Adatptive momentum transformer & $0.3$ & $0.1$ &  & $0.8$ \\
    \bottomrule
\end{tabular}
}
\end{center}
\end{table}

%\BW{[%Another concern about our work is that our improvement is limited. 
%We have done a more thorough search over the hyperparameters on the MNIST and machine translation task and observe more improvement on these tasks. 
%In particular, the bits/dim scores of momentum transformer, momentum transformer + momentum connection, and adaptive momentum transformer are 0.83, 0.80, and 0.78, which are significantly better than both softmax (0.84) and linear transformers (0.85). 
%Similarly, the BLUE scores of momentum transformer, momentum transformer + momentum connection, and adaptive momentum transformer on the IWSLT machine translation task are 22.25, 22.32, and 22.38, which are much better than linear transformers (21.37).
%]}

\subsection{Copy Task}

{We train momentum transformers and baseline models on a synthetic copy task to analyze their convergence speed.} In this task, the model has to duplicate a sequence of symbols. 
%as in \citep{katharopoulos2020transformers}. 
Each training and test sample has the form $0w0w$ where $w$ is a sequence of symbols collected from the set $\{1,\dots,N\}$. An example with the word $w$ of length 3 is given below.

\begin{center}
\begin{tabular}{|c|c|c|c|c|c|c|c|c|c|}
\hline
{\bf Example:} & 0 & 15 & 124 & 71 & 0  & 15  & 124  & 71  \\ \hline
\end{tabular}
\end{center}

In our experiments, we follow the same experimental setting as that used by \citet{katharopoulos2020transformers}. In particular, we use a sequence of maximum length 128 with 10 different symbols separated by a separator symbol. The baseline architecture for all methods is a 4-layer transformer with 8 attention heads and $D=32$. The models are trained with the RAdam optimizer %\citep{liu2020on} 
using a batch size of 64 and a learning rate of $10^{-3}$ which is reduced to $10^{-4}$ after 3000 iterations. Figure~\ref{fig:copy-convergence-analysis} shows the training loss and the test accuracy over epochs and over GPU time. Both the momentum and the adaptive momentum transformers converge much faster and achieve better training loss than the linear transformer. Notice that while the standard softmax transformer converges the fastest, it has quadratic complexity. 
%Also, note that the adaptive momentum transformer has similar performance as the momentum transformer without the need of tuning for the momentum value. %\BW{$d=32$ making $n$ to be the dominating factor.}

\subsection{Image Generation}
Transformers have shown great promise in autoregressive generation applications \citep{radford2019language,child2019generating}, such as autoregressive image generation \citep{ramesh2020dalle}. However, the training and sampling procedure using transformers are quite slow for these tasks due to the quadratic computational time complexity and the memory scaling with respect to the sequence length. In this section, we train our momentum-based transformers and the baselines with causal masking to predict images pixel by pixel and compare their performance. In particular, we demonstrate that, like linear transformers, both momentum and adaptive momentum transformers are able to generate images much faster than the standard softmax transformer. Furthermore, we show that momentum-based transformers converge much faster than linear transformers while achieving better bits per dimension (bits/dim). We compare the generated images by different models in the Appendix. Note that momentum and adaptive momentum transformers also generate images with constant memory per image like linear transformers.

%\subsubsection{4.2.1. MNIST}~~
\begin{figure}[t!]
\centering
\includegraphics[width=0.9\linewidth]{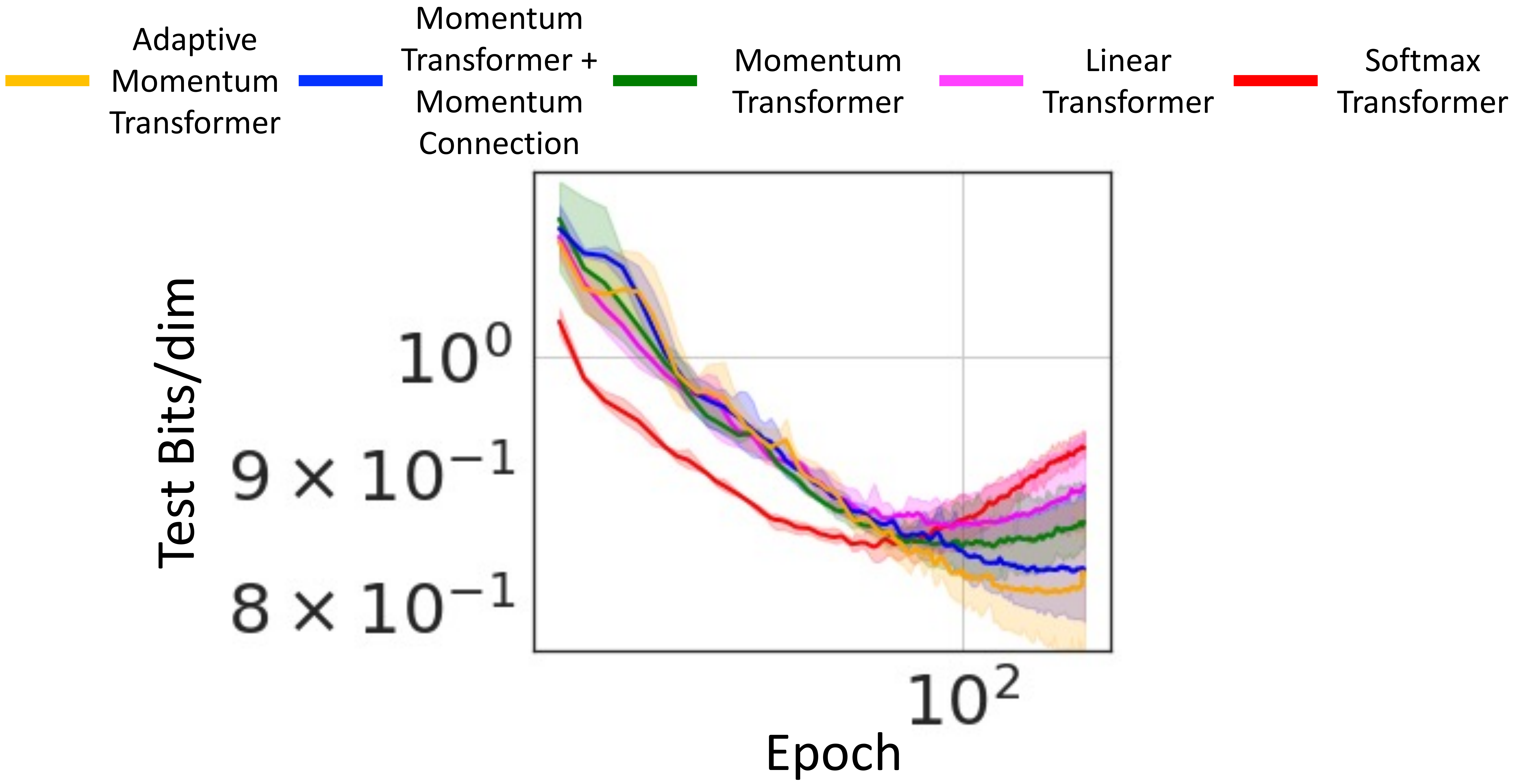}\vspace{-0.6cm}
\caption{Momentum transformers outperform linear transformers on the MNIST image generation task. Adaptive momentum transformer achieves the best test bits/dim.}
\label{fig:mnist-convergence-analysis}
\end{figure}

\bgroup
\renewcommand{\arraystretch}{1.0}
\begin{table*}[t!]
    \begin{center}
    \begin{tabular}{lcrl}
        Method & Bits/dim & \multicolumn{2}{c}{Images/sec} \\
        \hline
        Standard softmax transformer & 0.84 & 0.45 & (1$\times$) \\
        Linear transformer & 0.85 & 142.8 & (317$\times$)  \\
        \hline
        Momentum transformer & 0.84 & 139.7 & (310$\times$) \\
        Momentum transformer + momentum connection & 0.82 & 135.5 & (301$\times$) \\
        Adaptive momentum transformer & 0.80 & 134.9 & (300$\times$)  \\
    \end{tabular}
    \end{center}%\vspace{-0.8cm}
    \caption{Momentum transformers achieve better test bits/dim than both softmax and linear transformers on MNIST generation. %image generation task.
    }
    \label{tab:mnist}
\end{table*}
\egroup

\paragraph{MNIST.}
We first examine our momentum-based transformers on the MNIST image generation task. MNIST
is a popular benchmark dataset used for image recognition and generation. For all methods, we train a 8-layer transformer with 8 attention heads and the embedding size of 256, which corresponds to 32 dimensions per head. The feedforward dimensions are 4 times larger than the embedding size. A mixture of 10 logistics is used to model the output as in \citep{salimans2017pixelcnn}. For training, we use the RAdam optimizer with a learning rate of $10^{-4}$ and train all models for 250 epochs except for the adaptive momentum transformer. 

We report the bits/dim and image generation throughput in Table~\ref{tab:mnist}. Compared to the linear transformer, all momentum-based transformers not only attain better bits/dim but also have comparable image generation throughput, justifying the linear complexity of our models. In addition, we demonstrate that the adaptive momentum transformer converges much faster than the baseline models in Figure~\ref{fig:mnist-convergence-analysis}. Momentum-based transformers even outperform softmax transformers in this task.

\begin{figure*}[t!]
\centering
\includegraphics[width=0.9\linewidth]{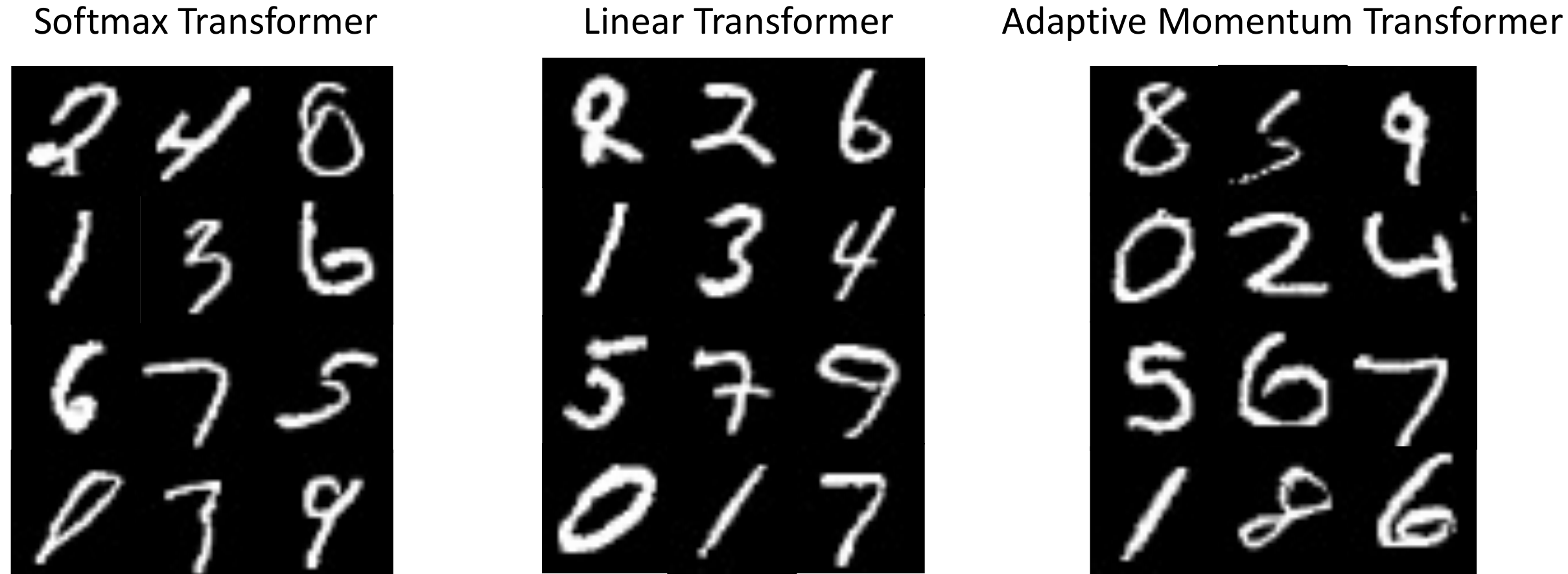}%\vspace{-0.5cm}
\caption{MNIST samples generated by the standard softmax transformer (left)~\citep{vaswani2017attention}, the linear transformer (middle)~\citep{katharopoulos2020transformers}, and the adaptive momentum transformer (right).}
\label{fig:mnist-sample}
\end{figure*}

We also compare our adaptive momentum transformer with the standard softmax and linear transformer qualitatively. In particular, we analyze the models trained for the MNIST image generation task and show the generated images from each model in Figure~\ref{fig:mnist-sample}. We observe that the quality of images generated from the adaptive momentum transformer and linear transformer is as high as the quality of images generated from the softmax transformer while the first two models are much more computational and memory efficient.

%\subsubsection{4.2.2. CIFAR10}
\paragraph{CIFAR10.}
Next, we investigate the advantages of our momentum-based transformers when the sequence length and the number of layers in the model increase. We consider the CIFAR-10 image generation task, in which we train 16-layer transformers to generate CIFAR-10 images. The configuration for each layer is the same as in the MNIST experiment. For the linear transformer and our momentum-based transformer, we use a batch size of 4 while using a batch size of 1 for the standard softmax transformer due to the memory limit of the largest GPU available to us, i.e., NVIDIA V100. This is similar to the setting in \citep{katharopoulos2020transformers}. Like in the MNIST image generation task, our momentum-based transformers outperform the linear transformer in terms of bits/dim while maintaining comparable image generation throughput. This is a very expensive task, limiting us to perform a thorough hyperparameter search; we believe better results can be obtained with a more thorough hyperparameter search.

\bgroup
\renewcommand{\arraystretch}{1.0}
\begin{table*}[t!]
    \begin{center}
    \begin{tabular}{lcrl}
        Method & Bits/dim & \multicolumn{2}{c}{Images/sec} \\
        \hline
        Standard softmax transformer & 3.20 & 0.004 & (1$\times$) \\
        Linear transformer & 3.44 & 17.85 & (4462$\times$)  \\
        \hline
        Momentum transformer & 3.43 & 17.52 & (4380$\times$) \\
        Momentum transformer + momentum connection & 3.41 & 17.11 & (4277$\times$) \\
        Adaptive momentum transformer & 3.38 & 17.07 & (4267$\times$)  \\
    \end{tabular}
    \end{center}%\vspace{-0.8cm}
    \caption{%Our 
    Momentum-based transformers achieve better test bits/dim than %the 
    linear transformer on CIFAR10 image generation task.}
    \label{tab:cifar10}
\end{table*}
\egroup

\begin{table*}[!t]
\centering
{\scriptsize
%\footnotesize
\begin{tabular}{c|c|c|c|c|c|c}
\hline
Model          & ListOps (2K) & Text (4K) & Retrieval (4K) & Image (1K) &  Pathfinder (1K) & Avg \\ \hline%\hline
Softmax & 37.10 (37.10) & 64.17 (65.02) & 80.71 (79.35) & 39.06 (38.20) & 72.48 (74.16) & \bf 58.70 (58.77) \\
\hline%\hline
Linear  & 18.30 & 64.22 & 81.37 & 38.29 & 71.17 & 54.67 \\
\hline%\hline
Performer  & 18.80 & 63.81 & 78.62 & 37.07 & 69.87 & 53.63 \\
\hline%\hline
Reformer  & 19.05 & 64.88 & 78.64 & 43.29 & 69.36 & 55.04 \\
\hline%\hline
Linformer  & \bf 37.25 & 55.91 & 79.37 & 37.84 & 67.60 & 55.59 \\
\hline%\hline
Momentum transformer & 19.56 & 64.35 & 81.95 & 39.40 & 73.12 & 55.68  \\
Adaptive momentum  & 20.16 & \bf 64.45 & \bf 82.07 & \bf 39.53 & \bf 74.00 & 56.04 \\
transformer &  &  &  &  &  & \\
\hline
\end{tabular}}
\hspace{0.1em}%\vspace{-0.4cm}
\caption{Results on the LRA tasks. We report the test classification accuracy for each task and average accuracy across all tasks. The momentum-based transformers, in particular, the adaptive momentum transformer, outperforms all other transformers except on the ListOps. 
%linear transformer in all tasks and attain better results than the standard transformer in most tasks except the Listops. 
The numbers in the parenthesis are from the paper \citep{xiong2021nystromformer}. Unit: \%. 
%\BW{[Need some discussion. In particular, we are better in most cases except ListOps there is a big loss, which was because of Linear transformer's problem.]}
}\label{tab:lra}
\end{table*}

\bgroup
\renewcommand{\arraystretch}{1.0}
\begin{table*}[t!]
    \begin{center}
    \begin{tabular}{lcc}
        Method & BLEU Score & Speed (tokens/s) \\
        \hline
        Standard softmax transformer & 24.34 & 5104 \\
        Linear transformer & 21.37 & 1382  \\
        \hline
        Momentum transformer & 22.11 & 1398 \\
        Momentum transformer + momentum connection & 22.14 & 1403 \\
        Adaptive momentum transformer & 22.20 & 1410 \\
    \end{tabular}
    \end{center}%\vspace{-0.7cm}
    \caption{BLEU scores and tokens per second from machine translation models trained on IWSLT
show the advantages of our momentum-based transformers.
The number of trainable parameters is almost the same for all models,
up to the small difference introduced by the momentum mechanism in our models. Momentum-based transformers outperform the linear transformer in generation quality in terms of BLEU score and obtain comparable generation efficiency in terms of tokens per second.}
    \label{tab:wikitext103}
\end{table*}
\egroup

\subsection{Long-Range Arena}
In this experiment, we evaluate our model on tasks that involve longer sequence lengths in the Long Range Arena (LRA) benchmark~\citep{tay2021long}. We show that the momentum-based transformer outperforms the baseline linear transformer and other popular efficient transformers, including performer~\citep{choromanski2021rethinking}, reformer~\citep{Kitaev2020Reformer}, and linformer~\citep{wang2020linformer}.
%in all tasks. 
We also demonstrate that our momentum-based transformer yields better accuracy than the standard softmax transformer~\citep{vaswani2017attention} in most tasks except the ListOps. These results justify the advantage of our momentum-based transformers in capturing long-term dependency. 

\paragraph{Datasets and metrics} We consider all five tasks in the LRA benchmark \citep{tay2021long}, 
including Listops, byte-level IMDb reviews text classification, byte-level document retrieval, CIFAR-10 image classification on sequences of pixels, and Pathfinder. These tasks involve long sequences of length $2K$, $4K$, $4K$, $1K$, and $1K$, respectively. We follow the setup/evaluation protocol in \citep{tay2021long} and report the test accuracy for each 
%individual 
task and the average result across all tasks.  
%Details on the LRA benchmark 
%Long Range Arena (LRA) Benchmark 
%are given in %the original paper
%\citep{tay2021long}. 

\paragraph{Models and training} All models have 2 layers, 64 embedding dimension, 128 hidden dimension, 2 attention heads. Mean pooling is applied in all models. Also, we use the nonlinear activation $elu(x) + 1$ for the linear transformer. Our implementation uses the public code by ~\citet{xiong2021nystromformer} as a starting point, and we follow their training procedures. The training setting and additional baseline model details are provided in the configuration file used in~\citep{xiong2021nystromformer} and can be found at \url{https://github.com/mlpen/Nystromformer/blob/main/LRA/code/lra_config.py}.

\paragraph{Results} We summarize our results in Table~\ref{tab:lra}. Both momentum-based transformers %and adaptive momentum transformer 
outperform linear transformers in all tasks and yield better accuracy than the standard softmax transformer in most tasks except the Listops. The adaptive momentum transformer performs the best on every task except the LipsOps, far behind the softmax transformer and Linformer.

\subsection{Non-Autoregressive Machine Translation}
%All of the above 
The previous experiments are for auto-regressive tasks. In this %last 
experiment, we demonstrate that the benefits of our momentum-based transformers also hold for a non-autoregressive task. We consider a machine translation task on the popular IWSLT' 16 En-De dataset. We follow the setting in~\citep{lee-etal-2018-deterministic}. In particular, we tokenize each sentence using a script from Moses~\citep{koehn-etal-2007-moses} and segment each word into subword units using BPE~\citep{sennrich-etal-2016-neural}. We also use $40K$ tokens from both source and target. Our baseline model is the small transformer-based network in~\citep{lee-etal-2018-deterministic} with $ d_{model}=278$, $d_{hidden}=507$, $ p_{dropout}=0.1$, $ n_{layer}=5$, and $ n_{head}=2$. This model has 5 layers, and each layer has 2 attention heads.  A depiction of this architecture is given in Figure 2 in~\citep{lee-etal-2018-deterministic}. The block ``Encoder" encodes the input X, the block ``Decoder 1" computes the conditional $\log p(Y^{0}|X)$, and the block ``Decoder 2" is shared across iterative refinement steps, calculating $\log p(Y^{\ell}|\hat{Y^{\ell -1}}, X)$. For the baseline standard softmax transformer model, we use the same architecture as in~\citep{lee-etal-2018-deterministic} with an additional positional attention and using the highway layer in the decoders. For the linear and our momentum-based transformer models, we replace the softmax attention with the linear attention and momentum-based attention, respectively. During training, we use linear annealing learning rate scheduling (from $3\times10^{-4}$ to $10^{-5}$). We do not use label smoothing nor average multiple check-pointed models.

Table~\ref{tab:wikitext103} reports the results in terms of generation quality, measured by the BLEU score~\citep{Papineni02bleu:a}, and generation efficiency, measured by the number of generated tokens per second. Consistent with other experiments above, our momentum-based transformers obtain better BLEU scores than the linear transformer in this non-autoregressive setting. Furthermore, in terms of generation efficiency, momentum-based models are comparable with the linear transformer and much more efficient than the standard softmax transformer.

\subsection{{Ablation Studies}}

\begin{figure}[!t]
\centering
%\vskip -0.2cm
\includegraphics[width=0.9\linewidth]{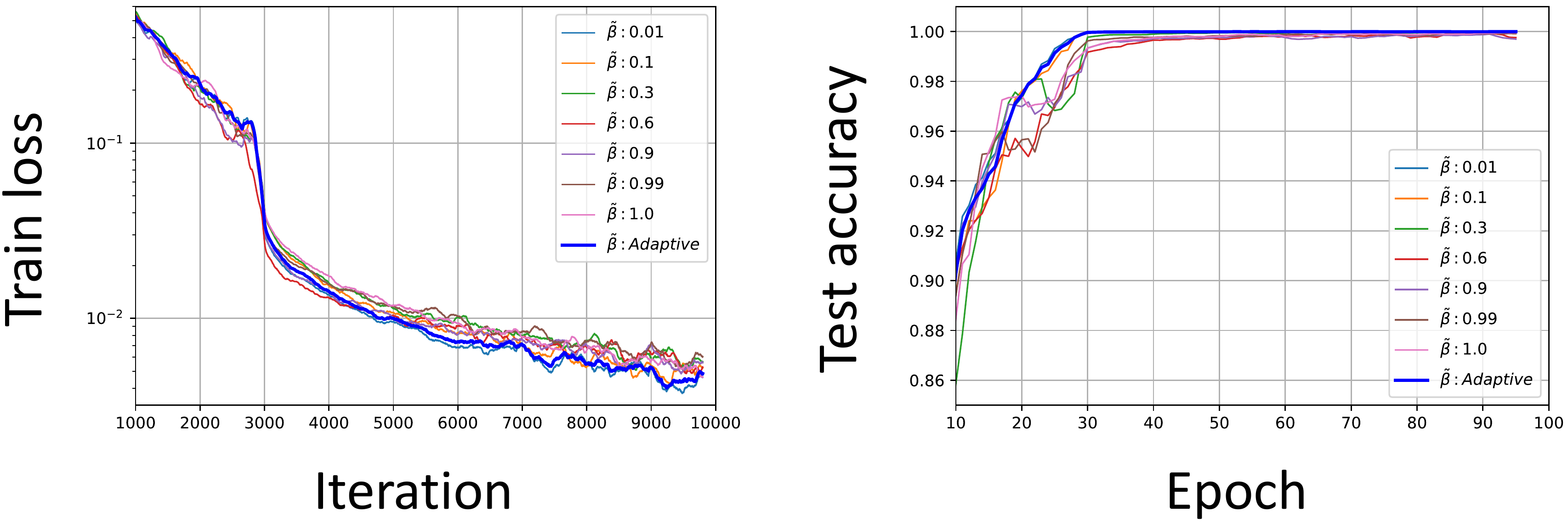}
\vskip -0.5cm
\caption{Ablation study of the effects of $\tilde{\beta}$ on the performance of momentum transformer with momentum connection in the synthetic copy task. We report the train loss (Left) and test accuracy (Right). Adaptive $\tilde{\beta}$ (the blue curve) yields similar train loss and test accuracy as the best constant $\tilde{\beta}$ found by a careful search.
%We use $N=256$ and $158$ hidden units for MNIST and TIMIT tasks, resp.
%Green denotes better results.
}
\label{fig:copy_rmu_ablation}
%\vspace{0.1in}
%\vskip -0.5cm
\end{figure}

\begin{figure}[!t]
\centering
%\vskip -0.2cm
\includegraphics[width=0.9\linewidth]{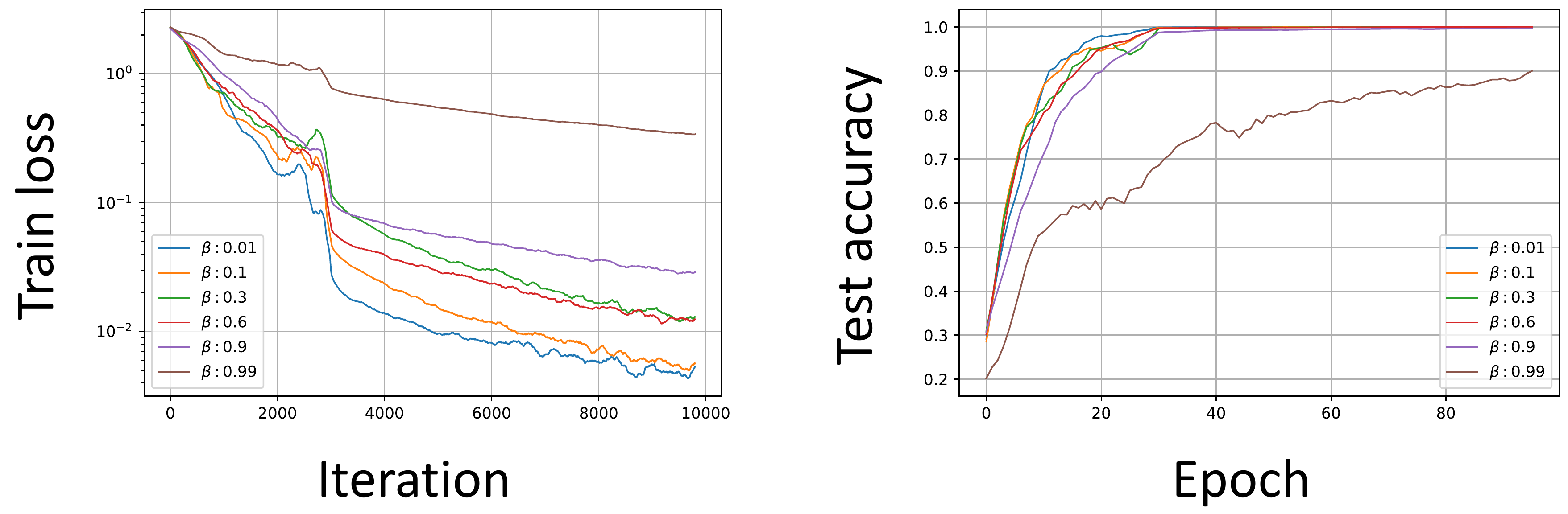}
\vskip -0.5cm
\caption{Ablation study of the effects of momentum $\beta$ in the momentum attention on the performance of adaptive momentum transformer in the synthetic copy task. We report the train loss (Left) and test accuracy (Right). Smaller $\beta$ yields better results.
%We use $N=256$ and $158$ hidden units for MNIST and TIMIT tasks, resp.
%Green denotes better results.
}
\label{fig:copy_rmu_ablation_beta}
%\vspace{0.1in}
%\vskip -0.5cm
\end{figure}

{ 
%{\bf \noindent Effect of the momentum $\tilde{\beta}$ in the momentum connection on the performance of momentum transformer with momentum connection.}
{\bf \noindent Effects of $\tilde{\beta}$ on the performance of momentum transformer with momentum connection.}
 We have conducted an ablation study to analyze how} { values of %the momentum 
 $\tilde{\beta}$  in momentum connection influence the the performance of momentum transformer with momentum connection in the synthetic copy task. Figure~\ref{fig:copy_rmu_ablation} shows that the transformer with adaptive $\tilde{\beta}$ (the blue curve) achieves as good train loss and test accuracy as the transformer with the best constant $\tilde{\beta}$ that are carefully fine-tuned, i.e. $\tilde{\beta}=0.01$. Note that our adaptive approach to computing $\tilde{\beta}$ eliminates the need of expensive search} {for the good values of $\tilde{\beta}$.\\}
 
\noindent { {\bf Effects 
%of momentum 
$\beta$ in the momentum attention on the performance adaptive momentum transformer.} We have conducted another ablation study on how the values of %the momentum 
$\beta$} {in the momentum attention affect the performance of the adaptive momentum transformer in the synthetic copy task. Figure~\ref{fig:copy_rmu_ablation_beta} demonstrates that smaller $\beta$ yields better results in terms of train loss and test accuracy than large ones. We also notice that when $\beta \ge 1$, the training is unstable and does not converge.}

%\BW{[1. We have added numerical results on momentum attention with the gating mechanism. 2. We have further compared with fast weight transformers. 
%Moreover, we have added the comparison with other transformers such as performers, linformers, and reformers that the reviewers suggested. 
%On the LRA benchmark, the average accuracies of our momentum transformers are better than the accuracies of these models reported in [61].
%]}

\section{Concluding Remarks}
\label{sec:conclusions}
In this paper, we developed a new class of efficient transformers, i.e., momentum transformers, which have the same memory and computational complexity as the recently developed linear transformer. We developed momentum transformers based on an analogy between the RNN formulation of causal linear attention and gradient descent. Then we integrate the momentum into causal linear attention following the heavy ball method. Furthermore, we introduce an additional momentum into the residual connection between the attention $\hat{\mV}$ and the input ${\mX}$ in~\eqref{eqn:res-connect} to further improve the performance of the model. To eliminate the computational overhead for tuning the momentum-related hyperparameters and enhancing momentum transformers' performance, we developed the adaptive momentum transformer that can adaptively compute the momentum values based on the optimal momentum choice for the heavy ball method for quadratic optimization. An interesting observation is that the momentum attention can be understood as a reweighting between the product of the ``key'' and ``value'' in the standard attention model. 
%The momentum transformer, especially the adaptive momentum transformer, performs remarkably better than the linear transformer on both autoregressive and non-autoregressive tasks. 
There are numerous avenues for future work: 1) can we develop momentum transformers based on other popular optimization algorithms beyond the heavy ball method, e.g., Adam?
And 2) can we design better weighting schemes to improve the performance of transformers?

% Acknowledgments---Will not appear in anonymized version
\clearpage

%\bibliography{references_aaai}

\end{document}